\documentclass[sigconf]{acmart}

\usepackage{algorithm}
\usepackage{algorithmic}
\usepackage{multirow}
\usepackage{subfigure}
\newtheorem{definition}{Definition}
\newtheorem{lemma}{Lemma}

\AtBeginDocument{%
  \providecommand\BibTeX{{%
    \normalfont B\kern-0.5em{\scshape i\kern-0.25em b}\kern-0.8em\TeX}}}

\setcopyright{acmcopyright}
\copyrightyear{2018}
\acmYear{2018}
\acmDOI{10.1145/1122445.1122456}

\acmConference[Woodstock '18]{Woodstock '18: ACM Symposium on Neural
  Gaze Detection}{June 03--05, 2018}{Woodstock, NY}
\acmBooktitle{Woodstock '18: ACM Symposium on Neural Gaze Detection,
  June 03--05, 2018, Woodstock, NY}
\acmPrice{15.00}
\acmISBN{978-1-4503-XXXX-X/18/06}

\begin{document}

\title{DNN2LR: Automatic Feature Crossing for Credit Scoring}


\author{Qiang Liu}
\email{qiangliu2018@tsinghua.edu.cn}
\affiliation{%
  \institution{Tsinghua University}
  \city{Beijing}
  \country{China}
}

\author{Zhaocheng Liu}
\email{zhaocheng.liu@realai.ai}
\affiliation{%
  \institution{RealAI}
  \city{Beijing}
  \country{China}
}

\author{Haoli Zhang}
\email{haoli.zhang@realai.ai}
\affiliation{%
  \institution{RealAI}
  \city{Beijing}
  \country{China}
}

\author{Yuntian Chen}
\email{chenyt01@pcl.ac.cn}
\affiliation{%
  \institution{Peng Cheng Laboratory}
  \city{Shenzhen}
  \country{China}
}

\author{Jun Zhu}
\email{dcszj@mail.tsinghua.edu.cn}
\affiliation{%
  \institution{Tsinghua University}
  \city{Beijing}
  \country{China}
}

\renewcommand{\shortauthors}{Liu, et al.}

\begin{abstract}
Credit scoring is a major application of machine learning for financial institutions to decide whether to approve or reject a credit loan.
For sake of reliability, it is necessary for credit scoring models to be both accurate and globally interpretable.
Simple classifiers, e.g., Logistic Regression (LR), are white-box models, but not powerful enough to model complex nonlinear interactions among features.
Fortunately, automatic feature crossing is a promising way to find cross features to make simple classifiers to be more accurate without heavy handcrafted feature engineering.
However, credit scoring is usually based on different aspects of users, and the data usually contains hundreds of feature fields.
This makes existing automatic feature crossing methods not efficient for credit scoring.
In this work, we find local piece-wise interpretations in Deep Neural Networks (DNNs) of a specific feature are usually inconsistent in different samples, which is caused by feature interactions in the hidden layers.
Accordingly, we can design an automatic feature crossing method to find feature interactions in DNN, and use them as cross features in LR.
We give definition of the interpretation inconsistency in DNN, based on which a novel feature crossing method for credit scoring prediction called DNN2LR is proposed.
Apparently, the final model, i.e., a LR model empowered with cross features, generated by DNN2LR is a white-box model.
Extensive experiments have been conducted on both public and business datasets from real-world credit scoring applications.
Experimental shows that, DNN2LR can outperform the DNN model, as well as several feature crossing methods.
Moreover, comparing with the state-of-the-art feature crossing methods, i.e., AutoCross, DNN2LR can accelerate the speed for feature crossing by about $10$ to $40$ times on datasets with large numbers of feature fields.
\end{abstract}





\maketitle

\section{Introduction}

Financial institutions, e.g., banks and online lending companies, evaluate the credit of users based on their profiles, loan history, repayment history and other behavior records.
They aim to predict the default probability of each user, so that they can decide whether to approve or reject a credit loan.
This is an important task due to the huge amount of loan applications \cite{babaev2019rnn,hu2020loan,west2000neural,bastos2007credit}.
For reliability, besides the accurate prediction of default, interpretability, which is usually a regulatory requirement, is also necessary \cite{babaev2019rnn}.

Some white-box models, e.g., Logistic Regression (LR), are simple and globally interpretable \cite{hall2019introduction}.
However, LR usually has relatively poor performances, and it is hard for LR to model complex nonlinear interactions among features.
To improve the performances of LR, heavy handcrafted feature engineering is usually required.
Meanwhile, black-box models such as Deep Neural Networks (DNNs) \cite{zhang2016deep,west2000neural} and tree ensemble models \cite{chen2016xgboost,ke2017lightgbm,bastos2007credit} are able to take use of complex nonlinear feature interactions.
Recently, these complex models show the possibility of being interpreted locally, which means assigning a piece of local interpretation weights for each sample \cite{li2015visualizing,ribeiro2016should,chu2018exact,lundberg2018consistent,mase2019explaining,aas2019explaining,guan2019towards}.
However, as pointed in \cite{guidotti2018survey,hall2019introduction}, different from LR, DNNs and tree ensemble models are still not globally interpretable, or can only be globally interpreted with an approximate linear importance vector.
In \cite{guidotti2018survey}, the global interpretability is formally defined as, \emph{we are able to understand the whole logic of a model and follow the entire reasoning leading to all the different possible outcomes}.
With only local interpretability for each sample, the credit scoring models can only assist the manual evaluation of loans, or analyze reasons after approval or rejection.
On the other hand, with global interpretability of credit scoring models, we can perform credit evaluation automatically, and understand the reasons of decisions in advance.
In a word, directly applying conventional classifiers is not optimal for credit scoring.

\begin{figure}
\centering
\includegraphics[width=0.45\textwidth]{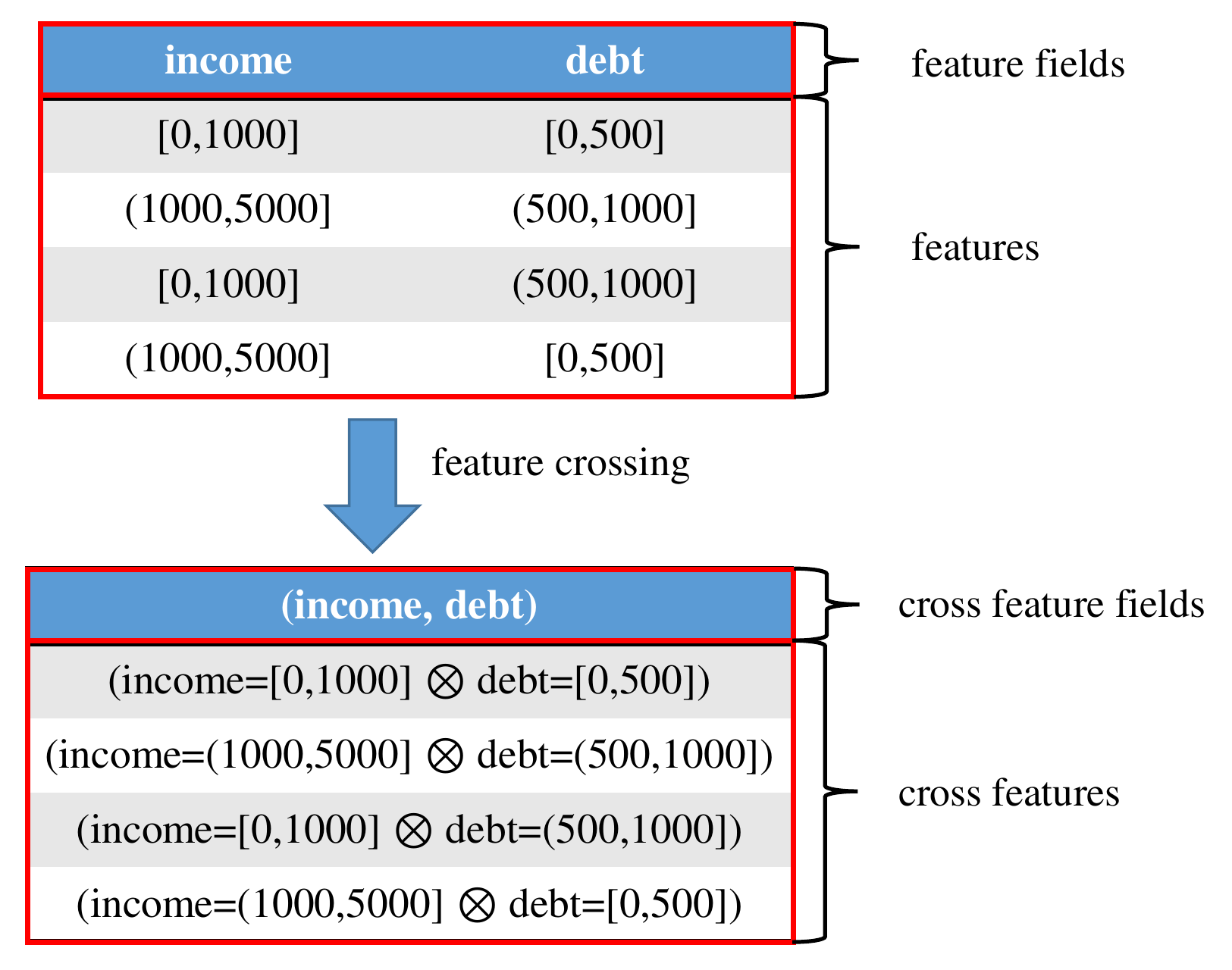}
\caption{An example of second-order feature crossing.}
\label{fig:example}
\end{figure}

Fortunately, automatic feature crossing is a promising way to capture nonlinear interactions among features \cite{chapelle2015simple,cheng2016wide,yuanfei2019autocross}.
As a practical direction in automated machine learning \cite{quanming2018taking}, it calculates second- or higher-order cross-product of categorical features.
For example, the cross feature representing a female teacher can be denoted as (``gender=female'' $\otimes$ ``occupation=teacher'').
Meanwhile, feature discretization has been proven useful to improve the capability of numerical features \cite{liu2002discretization,chapelle2015simple,liu2020an}.
Thus, we usually conduct feature discretization on numerical feature fields to generate corresponding categorical feature fields for automatic feature crossing \cite{yuanfei2019autocross}.
An example of second-order feature crossing can be found in Fig. \ref{fig:example}, in which we can obtain the cross feature field (income, debt) to measure a user's repayment ability.
As pointed in \cite{yuanfei2019autocross}, cross features, instead of latent embeddings or latent representations in DNN, are highly interpretable.
Via automatic generation of cross features, we can make the simple and globally interpretable LR model more accurate without heavy handcrafted feature engineering.
According to the analysis on Generalized Additive Model (GAM) \cite{lou2012intelligible} and GA$^2$M \cite{lou2013accurate}, the LR model empowered with cross features is a GAM with both second- and higher-order feature interactions, which is fortunately a white-box model.

Previous works on automatic feature crossing mostly try to explicitly search in the set of possible cross feature fields \cite{lou2013accurate,chapelle2015simple,katz2016explorekit,khurana2018feature,liu2020autofis}, including the state-of-the-art method AutoCross \cite{yuanfei2019autocross}.
In the application of credit scoring, we usually need to evaluate the user credit from different aspects.
This results in hundreds of feature fields in the data for credit scoring.
For example, there are $100$ to $300$ feature fields in several typical public credit scoring datasets, as well as in our real applications in credit scoring\footnote{Details are illustrated in Sec. \ref{app:data}}.
Thus, the candidate set of possible cross feature fields is inevitably large, which leads to the low efficiency of existing searching methods.
Suppose we have $100$ feature fields, there will be $4950$, $161700$ and $3921225$ second-, third- and forth-order cross feature fields respectively in the candidate set.
Meanwhile, for the task of Click-Through-Rate (CTR) prediction, various feature crossing modules are proposed based on deep learning \cite{qu2016product,cheng2016wide,wang2017deep,guo2017deepfm,lian2018xdeepfm,song2019autoint}.
Unfortunately, inherit from deep learning technologies, these deep learning-based methods are not globally interpretable, and hard to explicitly generate all the cross features captured in the models.

As mentioned above, DNN can be locally interpreted, via gradient backpropagation \cite{li2015visualizing,selvaraju2017grad} or feature perturbation \cite{fong2017interpretable,guan2019towards}.
These methods assign a piece of local interpretation for each sample, and interpret DNN as a combination of numbers of linear classifiers \cite{montufar2014number,ribeiro2016should,chu2018exact}.
In this work, we observe that, local interpretations of a specific feature are usually inconsistent in different samples.
We show and prove that, such inconsistency is caused by feature interactions occurred in the hidden layers of DNN.
Then, we give definition of the interpretation inconsistency in DNN, so that we can take advantage of the strong expressive ability of DNN to find cross features.
This can extremely reduce the size of candidate set, and improve the efficiency for finding cross features on credit scoring datasets with large numbers of feature fields.

Accordingly, we propose a novel \textbf{DNN2LR} method for automatic feature crossing for credit scoring.
Specifically, we first calculate local interpretations in DNN via the gradient backpropagation from the prediction layer to the input features \cite{selvaraju2017grad,alvarez2018towards,li2015visualizing,smilkov2017smoothgrad,yuan2019interpreting}.
Then, we define and calculate the interpretation inconsistency of each feature in each sample, based which most frequently occurring cross feature fields are extracted as candidate set.
Finally, we train a LR model with both original feature fields and candidate feature fields, and select final useful cross feature fields according to their contribution on the validation set.
With DNN2LR, we can obtain a LR model empowered with the final set of cross feature fields, which is a white-box model with high accuracy and global interpretability simultaneously.
We conduct experiments on both public and business datasets from real applications in credit scoring, with $100$ to $300$ original feature fields.
Experiments demonstrates that, our proposed DNN2LR method can outperform several conventional classifiers and several automatic feature crossing methods.
Comparing with the state-of-the-art feature crossing method, i.e., AutoCross \cite{yuanfei2019autocross}, DNN2LR reduces the running time for feature crossing by about $10\times$ to $40\times$.
In a word, DNN2LR helps us constructing accurate white-box models for the task of credit scoring in an efficient way.
The main contributions of this paper are summarized as follows:
\begin{itemize}
\item We give definition of the interpretation inconsistency in DNN. Then, we show that the interpretation inconsistency is caused by feature interactions in DNN, and can help us to find cross features.
\item we propose a novel DNN2LR method for automatic feature crossing for credit scoring. DNN2LR can generate an accurate and compact candidate set of cross feature fields, so that the searching can be efficiently done on credit scoring datasets with hundreds of original feature fields.
\item Experimental results on several real-world credit scoring datasets show that, DNN2LR outperforms existing methods in both effectiveness and efficiency. DNN2LR can produce accurate white-box models for the task of credit scoring in an efficient way.
\end{itemize}

\section{Related Works}

In this section, we review some work related to the task of automatic feature crossing for credit scoring.
First, we review some methods that explicitly search for cross features.
Then, we review some deep learning-based methods for CTR prediction.

\subsection{Cross Feature Searching Methods}

It is a direct way to explicitly search for useful cross feature fields in a candidate set.
However, the candidate set for searching is usually inevitably large, which leads to low efficiency of learning feature crossing, especially on credit scoring datasets with hundreds of input feature fields.
Most of these works focus on generating second-order cross features
\cite{lou2013accurate,chapelle2015simple,katz2016explorekit,nargesian2017learning,kaul2017autolearn}.
GA$^2$M \cite{lou2013accurate} extends GAM \cite{lou2012intelligible} with second-order cross features, which is hard to optimize when the candidate set is large.
In \cite{chapelle2015simple}, the authors try to generate and select second-order cross feature fields according to Conditional Mutual Information (CMI).
However, once the mutual information of an original feature field is high, the generated cross feature fields containing it will also have high conditional mutual information.
AutoLearn \cite{kaul2017autolearn} selects cross feature fields by using regularized regression models, where it is hard to learn a regression model for all the cross feature fields on a wide dataset.
Some works \cite{katz2016explorekit,nargesian2017learning} takes meta-learning into consideration for feature generation. However, the effectiveness of meta-learning in these methods remains a question and requires extremely large amount of data.
There are also some methods incorporating genetic algorithm \cite{tran2016genetic} and reinforcement learning \cite{khurana2018feature,chen2019neural} for finding feature combinations.
However, with genetic algorithm or reinforcement learning, we still have a large space to explore.
Moreover, as introduced in \cite{khurana2018feature}, it also requires large amount of data for the training of reinforcement learning.
To tackle with above problems, AutoCross \cite{yuanfei2019autocross} presents an approximate framework to search in the large candidate set of both second- and higher-order cross feature fields more efficiently.
AutoCross uniformly divides the whole dataset into at least $\sum\nolimits_{i = 0}^{\left\lceil {\mathop {\log }\nolimits_2 t} \right\rceil  - 1} {\mathop 2\nolimits^i }$ batches, where $t$ is the size of candidate set in AutoCross.
Then AutoCross iteratively trains a field-wise LR model on part of the data to validate the contribution of a cross feature field.
Though AutoCross achieves state-of-the-art performances, it still faces low searching efficiency on some wide datasets.
Moreover, when the candidate set is large, data for training the field-wise LR model of each candidate cross feature field will be too little to produce reliable results.
Therefore, AutoCross may face problems in both effectiveness and efficiency on credit scoring datasets with hundreds of feature fields.

Besides above methods explicitly searching for cross features, some works \cite{he2014practical,shi2020safe} utilize tree ensemble models \cite{chen2016xgboost,ke2017lightgbm} for generating cross features.
In \cite{he2014practical}, each tree in GBDT (Gradient Boosting Decision Tree) \cite{ke2017lightgbm} corresponds to a cross feature field, and each leaf node in a tree corresponds to a cross feature.
As trees in GBDT may be deep, the generated cross features may be with too high orders, so that they are hard for human to understand.
And cross features in the same cross feature field usually share different meanings.
Besides, Factorization Machine (FM) \cite{rendle2010factorization} has been a successful way to explicitly capture second-order feature interactions.
To overcome the problem that conventional FM can only model second-order feature interactions Higher-Order FM (HOFM) \cite{blondel2016higher} and Interaction Machine (IM) \cite{yu2020deep} propose to model higher-order feature interactions in the FM architecture in linear time.
The calculation of feature interactions in FM is somehow product-based similarity.
However, it is hard for these factorization-based methods to capture all kinds of feature interactions in various real-world application scenarios.
Another drawback of FM is that, it models all feature interactions of specific-order, most of which are not useful for making predictions.
To deal with this problem, Automatic Feature Interaction Selection (AutoFIS) \cite{liu2020autofis} directly learns the weight for each cross feature field, for both second-order and higher-order, and thus an AutoFM model is obtained.
It promotes the performances of FM, but faces problems in high computational cost and optimization difficulty, when it is performed on credit scoring datasets with hundreds of feature fields and the desired interaction order is high.

\subsection{Deep Learning-based CTR Prediction Methods} \label{sec:DL}

Nowadays, some works try to design various deep learning-based crossing modules for the task of CTR prediction.
For better performances, these crossing modules are usually applied along with multi-layer fully-connected DNNs.
The Wide \& Deep model \cite{cheng2016wide} directly learns parameters of manually designed cross features in the wide component.
The Product-based Neural Network (PNN) \cite{qu2016product} applies inner-product or outer-product to capture second-order features.
In Deep \& Cross Network (DCN) \cite{wang2017deep}, the authors design an incremental crossing module, named CrossNet, to capture second-order as well as higher-order features.
Factorization-based methods have also been extended with deep architectures \cite{guo2017deepfm,yu2020deep,liu2020autofis}.
In xDeepFM \cite{lian2018xdeepfm}, Compressed Interaction Network (CIN) is proposed as a crossing module based on outer-product calculation of features.
AutoINT \cite{song2019autoint} is a combination of residual connections \cite{he2016deep} and a crossing module based on Self-Attention \cite{vaswani2017attention}.
With the residual connections, the fully-connected DNN is performed with the input of original features in AutoINT.
Convolutional neural networks are also incorporated for modeling local interactions among nearby features \cite{liu2015convolutional,liu2019feature}.
Moreover, feature interactions are modeled as graphs, and then graph neural networks are performed \cite{xie2020interactive,su2021detecting}.
Neural architecture search approaches have also been incorporated for finding feature interaction architectures in deep learning-based CTR prediction models \cite{song2020towards,khawar2020autofeature}.

Inherit from deep learning technologies, above feature crossing modules are mostly associated with hidden projection, in each layer or between adjacent layers.
This makes deep learning-based methods still black-box models and hard to be globally interpreted, which does not meet our requirements for credit scoring.

\begin{figure*}
\centering
\includegraphics[width=0.75\textwidth]{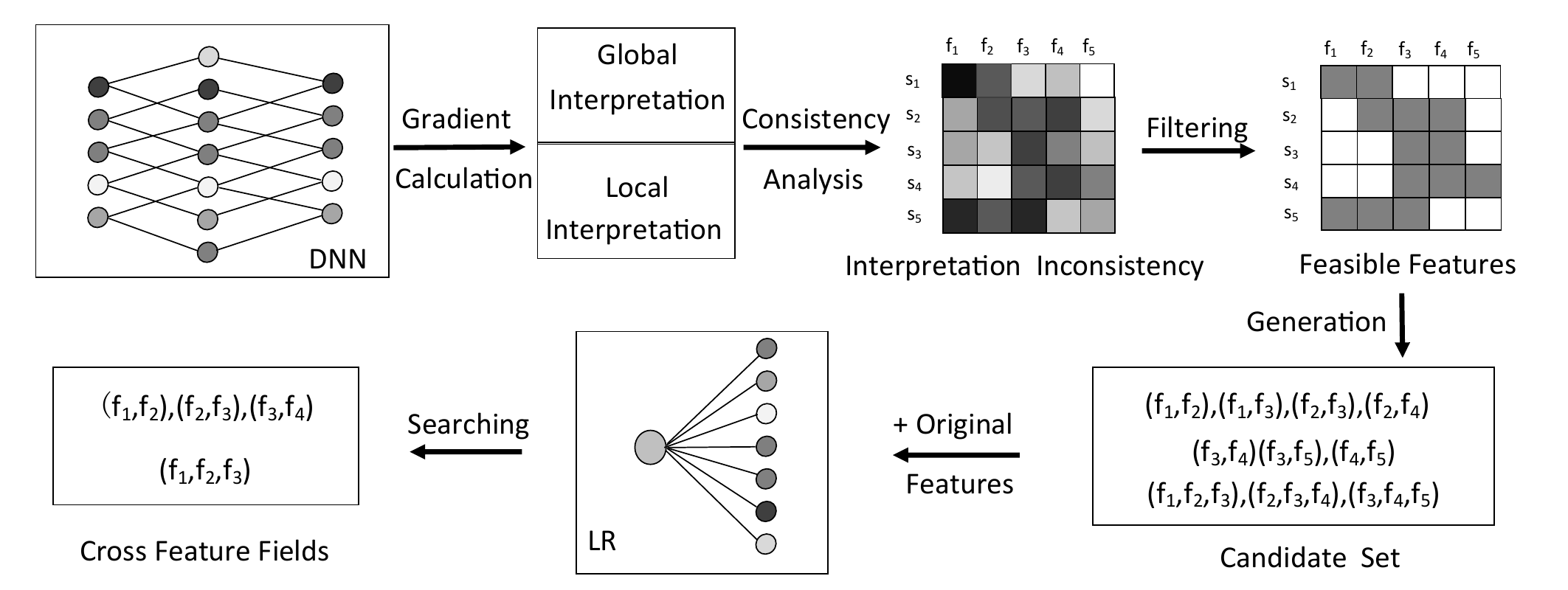}
\caption{Overview of DNN2LR, where $\mathop s\nolimits_i$ and $\mathop f\nolimits_j$ indicate the i-th sample and the j-th original feature field respectively.}
\label{fig:dnn2lr}
\end{figure*}

\section{Learning from Interpretations in DNN}

In this section, we present that we can learn feature crossing from interpretations in DNN for credit scoring.

\subsection{Local Interpretations}

DNN has been a powerful black-box model \cite{guidotti2018survey}.
Recently, extensive works have been done to study local piece-wise interpretability of DNN, which means assigning a piece of local interpretation weights for each sample, a DNN model can be regarded as a combination of numbers of linear classifiers \cite{montufar2014number,ribeiro2016should,lundberg2017unified,chu2018exact}.
Some works investigate the gradients from the final predictions to the input features in deep models, which can be applied in the visualization of deep vision models \cite{zhou2016learning,selvaraju2017grad,smilkov2017smoothgrad,alvarez2018towards}, as well as the interpretation of language models \cite{li2015visualizing,yuan2019interpreting}.
Perturbation on input features is also utilized to find local interpretations of both vision models \cite{fong2017interpretable} and language models \cite{guan2019towards}.
In this work, we adopt gradient backpropagation for the calculation of local interpretations.
Therefore, we can first define the local interpretations in DNN.
As introduced, we usually conduct feature discretization for numerical features.
Thus, we focus on analyzing DNN with embeddings of categorical features as input.

\begin{definition}(Local Interpretation) \label{def:local_interpretation}
Given a specific feature $\mathop x\nolimits_{k,f}$ associated with the feature field $\mathop f$ in a specific sample $k$, the corresponding local interpretation is
\begin{equation} \label{equation:local_interpretation}
\mathop I^l\nolimits_{k,f}  = \mathop w\nolimits_{k,f} \mathop e\nolimits_{k,f}^\top,
\end{equation}
where $\mathop e\nolimits_{\mathop k,f}$ denotes the corresponding feature embeddings of $\mathop x\nolimits_{k,f}$ in the DNN model, and $\mathop w\nolimits_{ k,f}$ is the local weights computed via gradient backpropagation
\begin{equation} \label{equation:local_weights}
\mathop w\nolimits_{k,f}  = \frac{{\partial \mathop {\hat y}\nolimits_k }}{{\partial \mathop e\nolimits_{k,f} }},
\end{equation}
where ${\mathop {\hat y}\nolimits_k }$ denotes the prediction made by the DNN model for the sample $k$.
\end{definition}

\subsection{Interpretation Inconsistency}

Local interpretation refers to the contribution of the corresponding feature to the final prediction in the corresponding sample.
Sometimes, local interpretations of a specific feature are inconsistent in different samples.
When local interpretations are consistent in different samples, it means the corresponding feature contributes to the final prediction on its own.
When local interpretations are inconsistent, it means the contribution of the corresponding feature is affected by other features.
The inconsistency of local interpretations in DNN is caused by nonlinear feature interactions in the hidden layers of DNN.

\begin{lemma}(Inconsistent Local Interpretations) \label{lemma:inconsistency}
If a feature field nonlinearly interacts with other feature fields in the hidden layers of DNN, features associated with the feature field will have inconsistent local interpretations among different samples.
\end{lemma}

Suppose feature fields $\mathop f\nolimits_1 ,\mathop f\nolimits_2 ,......$ are nonlinearly interacted in DNN, i.e., ${{\hat y}_k} = P\left( {{\rm{ }}{e_{k,{\rm{ }}{f_1}}},{\rm{ }}{e_{k,{\rm{ }}{f_2}}},......} \right)$, where $P\left(  \cdot  \right)$ is a nonlinear interaction function.
For example, $\mathop {\hat y}\nolimits_k = \mathop e\nolimits_{k,\mathop f\nolimits_1 } \mathop e\nolimits_{k,\mathop f\nolimits_2 }^ \top$.
Then, we can have the following local interpretation
\begin{equation}
{\rm{ }}I_{k,{\rm{ }}{f_i}}^l = \frac{{\partial P\left( {{\rm{ }}{e_{k,{\rm{ }}{f_1}}},{\rm{ }}{e_{k,{\rm{ }}{f_2}}},......} \right)}}{{{\rm{ }}\partial {e_{k,{\rm{ }}{f_i}}}}}{\rm{ }}e_{k,{\rm{ }}{f_i}}^ \top.
\end{equation}
In the example of $\mathop {\hat y}\nolimits_k = \mathop e\nolimits_{k,\mathop f\nolimits_1 } \mathop e\nolimits_{k,\mathop f\nolimits_2 }^ \top$, the local interpretations are $\mathop I\nolimits_{k,\mathop f\nolimits_1 }^l  = \mathop e\nolimits_{k,\mathop f\nolimits_2 } \mathop e\nolimits_{k,\mathop f\nolimits_1 }^ \top$ and $\mathop I\nolimits_{k,\mathop f\nolimits_2 }^l  = \mathop e\nolimits_{k,\mathop f\nolimits_1 } \mathop e\nolimits_{k,\mathop f\nolimits_2 }^ \top$.
Obviously, the local interpretation of ${\mathop x\nolimits_{k,\mathop f\nolimits_i } }$ is affected by the features associated with other feature fields.
As the values of ${x_{k,{\rm{ }}{f_{i'}}}}(i' \ne i)$ change among different samples, the local interpretations of ${\mathop x\nolimits_{k,\mathop f\nolimits_i } }$ are inconsistent.
In contrast, if there is no nonlinear interaction among $\mathop f\nolimits_1 ,\mathop f\nolimits_2 ,......$, we have an addition form, i.e., ${{\hat y}_k} = {\rm{ }}{P_1}\left( {{\rm{ }}{e_{k,{\rm{ }}{f_1}}}} \right) + {\rm{ }}{P_2}\left( {{\rm{ }}{e_{k,{\rm{ }}{f_2}}}} \right) + ......$, where $\mathop P\nolimits_i \left(  \cdot  \right)$ is an arbitrary function.
Then, we can have the following local interpretation
\begin{equation}
{\rm{ }}I_{k,{\rm{ }}{f_i}}^l = \frac{{\partial {P_i}\left( {{\rm{ }}{e_{k,{\rm{ }}{f_i}}}} \right)}}{{{\rm{ }}\partial {e_{k,{\rm{ }}{f_i}}}}}{\rm{ }}e_{k,{\rm{ }}{f_i}}^ \top.
\end{equation}
Obviously, with the same feature ${{x_{k,{\rm{ }}{f_i}}}}$, there will be consistent local interpretations among different samples.

To measure the degree of inconsistency among local interpretations of a specific feature in different samples, we first need to calculate its global interpretation, and then calculate the interpretation inconsistency between the local interpretation and the global interpretation.

\begin{definition}(Global Interpretation) \label{def:global_interpretation}
Given the feature $\mathop x\nolimits_{\mathop k,f}$, the corresponding global interpretation is
\begin{equation} \label{equation:global_interpretation}
\mathop I^g\nolimits_{k,f}  = \mathop {\bar w}\nolimits_{k,f} \mathop e\nolimits_{k,f}^\top,
\end{equation}
where ${\mathop {\bar w}\nolimits_{\mathop f } }$ is the average local weights of the feature $\mathop x\nolimits_{\mathop k,f}$ in all samples, named as global weights and formulated as
\begin{equation} \label{equation:global_weights}
\small
{{\bar w}_{k,f}} = \frac{1}{{\left\| {\left\{ {\mathop x\nolimits_{k',f}  = \mathop x\nolimits_{k,f} |k' \in \Omega } \right\}} \right\|}}\sum\limits_{k' \in \Omega ,\mathop x\nolimits_{k',f}  = \mathop x\nolimits_{k,f} } {{\rm{ }}{w_{k',f}}},
\end{equation}
where $\Omega$ is the set of samples.
\end{definition}

\begin{definition}(Interpretation Inconsistency) \label{def:interpretation_inconsistency}
Given a specific feature $\mathop x\nolimits_{\mathop k,f}$ in a specific sample $k$, the corresponding interpretation inconsistency is
\begin{equation} \label{equation:interpretation_inconsistency}
{\rm{ }}{d_{k,f}} = \mathop {\left\| {\left( {{\rm{ }}{w_{k,f}} - {{\bar w}_{k,f}}} \right){\rm{ }}e_{k,f}^ \top } \right\|}\nolimits_{\rm{2}}^{\rm{2}}.
\end{equation}
\end{definition}

According to Lemma \ref{lemma:inconsistency}, the interpretation inconsistency in DNN is able to lead us to generate a compact and accurate candidate set of cross feature fields.
And larger the values of interpretation inconsistency of a specific feature, more the corresponding feature field can work for feature crossing.
To verify the validity of interpretation inconsistency for feature crossing, we conduct empirical experiments in Sec. \ref{sec:vis}.

\section{DNN2LR}

In this section, we formally propose the DNN2LR approach for credit scoring.
DNN2LR consists of two steps: (1) generating a compact and accurate candidate set of cross feature fields; (2) searching in the candidate set for the final cross feature fields.
Fig. \ref{fig:dnn2lr} provides an overview of the proposed DNN2LR approach.

\subsection{Candidate Set Generation}

For generating the compact and accurate candidate set of cross features, we need to obtain the local piece-wise interpretations in DNN.
Thus, we need first to train a DNN model.
Specifically, in this work, we train a multi-layer fully-connected DNN.
As the original features are sparse categorical features, we use an embedding layer \cite{zhang2016deep} to transform the input features into low dimensional dense representations.
Then, the dense representations are passed through some linear transformation and nonlinear activation to obtain the predictions of samples, where we use ReLu as the activation for hidden layers, and sigmoid for the output layer to support binary classification tasks.
To be noted, though we focus on learning feature crossing from multi-layer fully-connected DNN in this work, our proposed approach can also works with other types of DNN, such as AutoINT \cite{song2019autoint} and xDeepFM \cite{lian2018xdeepfm}.

Based on the trained DNN model, in each validation sample $k$, we compute the interpretation inconsistency $\mathop d\nolimits_{k,f}$ of the feature $\mathop x\nolimits_{\mathop k ,f}$, as defined in Def. \ref{def:interpretation_inconsistency}.
Therefore, we obtain an interpretation inconsistency matrix $D$, where $\mathop D \left[ k,f \right]$ is the interpretation inconsistency value of the $f$-th feature field in the $k$-th sample.
Then, we conduct an element-wise filtering on matrix $D$ with a threshold $\eta$, which can be formulated as
\begin{equation}
    \mathop D \nolimits^{\rm{*}} \left[ k, f \right] =
    \begin{cases}
        1,& \mathop D \left[ k, f \right] \geq Quantile \left( D, 1 - \eta \right) \\
        0,& otherwise
    \end{cases}
    ,
\end{equation}
where $\mathop D^*$ is a binary feasible feature matrix, and $\mathop D \nolimits^{\rm{*}} \left[ k, f \right]$ indicates whether the $f$-th feature in the $k$-th sample has interacted with other features in the hidden layers.
$Quantile \left( D, 1 - \eta \right)$ denotes the $1 - \eta$ quantile of the matrix $D$, which means keeping the top $\eta$ elements ($0\% < \eta < 100\%$) in $D$ with largest values of interpretation inconsistency.
Then, for each feasible feature $\mathop D \nolimits^{\rm{*}} \left[ k, f \right] = 1$, the corresponding feature field $f$ can be used to generate candidate cross feature fields.

Finally, we greedily generate the candidate set of cross features fields.
Considering that extremely high-order cross features are rarely useful, as in \cite{yuanfei2019autocross}, we construct our candidate set with only second-order, third-order and forth-order cross feature fields.
To construct a compact and accurate candidate set, we need to find cross feature fields which are most frequently interacted in DNN.
According to the feasible feature matrix $\mathop D^*$, we count the occurrences of possible cross feature fields.
We then rank the cross feature fields in a descending order according to the corresponding occurrence frequency, and select the top $\varepsilon$ cross feature fields as our candidate set.
According to our experiments in Sec. \ref{sec:params}, we can use $\varepsilon = 3n$ as the optimal hyper-parameter, where $n$ is the size of the original feature fields.
Accordingly, a compact candidate set of cross feature fields is generated, and searching for useful cross feature fields can be efficiently conducted.

\begin{algorithm}[t]
    \caption{Searching for Final Cross Feature Fields.}
    \label{alg:searching}
    \small
    \begin{algorithmic}[1]
        \REQUIRE original feature field set $\mathop F = \{\mathop f_1, \mathop f_2, ...,$ $\mathop f_{n}\}$, candidate feature field set $\mathop C = \{\mathop c\nolimits_1, \mathop c\nolimits_2, ..., \mathop c\nolimits_{\varepsilon}\}$, validation set $\mathop \Omega \nolimits_{{\rm{valid}}}$, the corresponding labels $\mathop Y \nolimits_{{\rm{valid}}}$, sigmoid function $\gamma(\cdot)$, AUC computing function $\delta \left(  \cdot  \right)$, the bias term $b$ and lookup function $\rm{LOOKUP}(\cdot)$ in the sparse LR model.
        \ENSURE final set of cross feature fields $\mathop S\nolimits^*$.
        \STATE $\mathop S\nolimits^* = \{\}$;
        \STATE $\mathop E\nolimits_{{\rm{origin}}}  = {\rm{LOOKUP}}\left( {\left\{ {\mathop x\nolimits_{k,\mathop f\nolimits_i } |k \in \mathop \Omega \nolimits_{{\rm{valid}}} ,1 \le i \le n} \right\}} \right)$;
        \STATE $\mathop E\nolimits_{{\rm{cross}}}  = {\rm{LOOKUP}}\left( {\left\{ {\mathop x\nolimits_{k,\mathop c\nolimits_i } |k \in \mathop \Omega \nolimits_{{\rm{valid}}} ,1 \le i \le \varepsilon } \right\}} \right)$;
        \STATE ${\rm{logit}}\left( 0 \right) = \sum\limits_{1 \le i \le n} {\mathop E\nolimits_{{\rm{origin}}} \left[ {:,i} \right]}  + b$;
        \STATE ${\rm{AUC}}\left( 0 \right) = \delta \left( {\mathop Y\nolimits_{{\rm{valid}}} ,\gamma \left( {{\rm{logit}}\left( 0 \right)} \right)} \right)$;
        \FOR{$i$ in $[1, \varepsilon]$}
            \FOR{$j$ in $[1, \varepsilon]$}
                \IF{$c_j$ not in $\mathop S\nolimits^*$}
                    \STATE ${\rm{logit}}\left( j \right) = {\rm{logit}}\left( 0 \right) + \mathop E\nolimits_{{\rm{cross}}} \left[ {:,j} \right]$;
                    \STATE ${\rm{AUC}}\left( j \right) = \delta \left( {\mathop Y\nolimits_{{\rm{valid}}} ,\gamma \left( {{\rm{logit}}\left( j \right)} \right)} \right)$;
                \ENDIF
            \ENDFOR
            \STATE $k = \mathop {{\rm{argmax}}}_j \rm{AUC}(j)$;
            \IF{${\rm{AUC}}\left( k \right) > {\rm{AUC}}\left( 0 \right)$}
                \STATE $c_k \to \mathop S\nolimits^*$;
                \STATE ${\rm{logit}}\left( 0 \right) = {\rm{logit}}\left( k \right)$;
                \STATE ${\rm{AUC}}\left( 0 \right) = {\rm{AUC}}\left( k \right)$;
            \ELSE
                \STATE break;
            \ENDIF
        \ENDFOR
        \RETURN $\mathop S\nolimits^*$.
    \end{algorithmic}
\end{algorithm}

\subsection{Searching for Final Cross Feature Fields} \label{sec:searching}

After obtaining the candidate set of cross feature fields $\mathop C = \{\mathop c_1, \mathop c_2, ...,$ $\mathop c_{\varepsilon}\}$, we can search for final useful cross feature fields.
The searching strategy in AutoCross \cite{yuanfei2019autocross} has been proven effective.
Facing a large candidate set, AutoCross trains a LR model for each candidate cross feature field, and select final cross feature fields based on their contribution measured on the validation set.
Obviously, there are too many LR models to train, and the efficiency is low.
In contrast, we have a compact and accurate candidate set, and are able to feed all candidate cross feature fields into a single LR model.
Thus, we do not need to involve the complex searching structure in \cite{yuanfei2019autocross}, and can simply select useful cross feature fields during validation.

To search for useful cross feature fields, we need to train a sparse LR model consisting of both original feature fields $\mathop F = \{\mathop f_1, \mathop f_2, ...,$ $\mathop f_{n}\}$ and candidate cross feature fields $\mathop C = \{\mathop c_1, \mathop c_2, ...,$ $\mathop c_{\varepsilon}\}$.
For a specific sample $k$, the prediction can be made by sparse LR as
\begin{equation} \label{eq:sparse_LR}
\small
\mathop {\hat y}\nolimits_k  = \gamma \left( {\sum\limits_{1 \le i \le n} {{\rm{LOOKUP}}\left( {\mathop x\nolimits_{k,\mathop f\nolimits_i } } \right)}  + \sum\limits_{1 \le i \le \varepsilon } {{\rm{LOOKUP}}\left( {\mathop x\nolimits_{k,\mathop c\nolimits_i } } \right)}  + b} \right),
\end{equation}
where ${\mathop x\nolimits_{k,\mathop c\nolimits_i } }$ is the cross feature associated with cross feature field $c_i$ in the sample $k$, ${\rm{LOOKUP}}\left(  \cdot  \right)$ is a lookup function for the weight in sparse LR of the corresponding feature, $\gamma(\cdot)$ is the sigmoid function, and $b$ is a bias term.
Specifically, we first train the sparse LR model with only original features on the training set $\mathop \Omega \nolimits_{{\rm{train}}}$.
Then, we fix the weights of original features, and train the model with candidate cross features on the training set $\mathop \Omega \nolimits_{{\rm{train}}}$.
According to our experiments in Sec. \ref{sec:params}, we can use $\varepsilon = 3n$ as the optimal hyper-parameter.
Therefore, the time cost of training the above model is in the same order of magnitude with training with original features.

Based on the trained LR model, we conduct searching procedure for useful cross feature fields according to their contribution measured on the validation set $\mathop \Omega \nolimits_{{\rm{valid}}}$ according to the AUC (Area Under the ROC Curve) metric.
Pseudocode of the searching procedure is presented in Alg. \ref{alg:searching}.
Moreover, the steps 7-12 in Alg. \ref{alg:searching} are paralleled with multi-threading implementation, where the measuring of each candidate cross feature field is conducted on one thread.
The main idea in Alg. \ref{alg:searching} is that, based on parameters in the trained LR model, we measure the contribution in AUC of each candidate cross feature field on the validation set, and iteratively select those have positive contribution as the final set $\mathop S\nolimits^*$ of cross feature fields.
Moreover, we can further incorporate beam search \cite{cohen2019empirical} in Alg. \ref{alg:searching}.
And this will be further investigated in our experiments in Sec. \ref{sec:performance}.

\subsection{Deploying to Online Production}

After obtaining the final set of cross feature fields and a white-box model, we need to deploy the final model to online production.
As the final model is a LR model, and LR is widely-used for credit scoring or default prediction in banks or online lending companies, the final white-box model is easily deployed to the online system.
We only need to link the final LR model to the existing online LR serving system, and generate cross features based on data flows.

\section{Experiments}

In this section, we empirically evaluate our proposed DNN2LR approach.
We first describe settings of the experiments, then report and analyze the experimental results.

\begin{table}[t]
  \centering
  \small
\caption{Details of experimental datasets.}\label{tab:data}
    \begin{tabular}{c|ccc|cc}
    \toprule
    \multirow{2}[2]{*}{dataset} & \multicolumn{3}{c|}{\#samples} & \multicolumn{2}{c}{\#feature fields} \\
          & training & validation & testing & \#Num. & \#Cate. \\
    \hline
    BNP   & 73,165 & 18,291 & 22,865 & 109   & 23 \\
    PPD   & 36,000     & 9,000     & 15,000     & 218     & 109 \\
    Bussiness1 & 186,498 & 46,625 & 58,282 & 178   & 15 \\
    Bussiness2 & 120,989 & 30,247 & 52,169 & 302   & 56 \\
    \bottomrule
    \end{tabular}%
  \label{tab:data}%
\end{table}

\subsection{Datasets} \label{app:data}

We conduct automatic feature crossing experiments on $4$ datasets: BNP, PPD, Bussiness1 and Bussiness2.
BNP\footnote{\url{https://www.kaggle.com/c/bnp-paribas-cardif-claims-management/data}} is a public credit scoring dataset from a bank, and PPD\footnote{\url{https://www.kesci.com/home/competition/      56cd5f02b89b5bd026cb39c9/content/1}} is a public default prediction dataset from an online lending company.
Bussiness1 and Bussiness2 are business datasets from our real-world applications in credit scoring of banks, after anonymization and sanitization.
For BNP and PPD, we randomly use $75\%$ and $25\%$ samples for training and testing respectively.
For all the four datasets, we randomly use $20\%$ samples in the training set for validation.
Details of these datasets can be found in Tab. \ref{tab:data}.
There are about $100$ to $300$ feature fields in these datasets, characterizing users in profiles, loan history, repayment history and other behavior records.

\subsection{Settings} \label{app:setting}

In our experiments, we are going to verify whether we can obtain a white-box model with high accuracy and global interpretability simultaneously.
Specifically, we are going to verify whether DNN2LR can empower the simple LR model to achieve better performances comparing with conventional classifiers, as well as other competitive feature crossing methods.
First, we compared DNN2LR with several conventional common-used classifiers: \textbf{LR}, \textbf{DNN} and \textbf{GBDT}.
Then, some competitive feature crossing methods are included: \textbf{FM} \cite{rendle2010factorization}, \textbf{HOFM} \cite{blondel2016higher}, \textbf{AutoFM} \cite{liu2020autofis}, \textbf{GA$^2$M} \cite{lou2013accurate}, \textbf{GBDT+LR} \cite{he2014practical} and \textbf{AutoCross+LR} \cite{yuanfei2019autocross}.
Meanwhile, we further include several deep-learning based feature crossing modules: \textbf{CrossNet} \cite{wang2017deep}, \textbf{CIN} \cite{lian2018xdeepfm} and \textbf{Self-Attention} \cite{song2019autoint,vaswani2017attention}.
They are the feature crossing modules used in CTR prediction models DCN \cite{wang2017deep}, xDeepFM \cite{lian2018xdeepfm} and AutoINT \cite{song2019autoint} respectively.
To be noted, as discussed in Sec. \ref{sec:DL}, these deep learning-based methods are still black-box models, and hard to be globally interpreted.
Moreover, we run our proposed \textbf{DNN2LR} method $10$ times with different parameter initialization, where average results and significance test are reported.
The detailed settings of above compared methods are introduced as following.

\begin{table*}[t]
  \centering
  \small
  \caption{Performance comparison in terms of AUC (\%) and KS (\%). Larger the values, better the performances. $*$ denotes statistically significant improvement, measured by t-test with p-value$<0.01$, over the second best method on each dataset.}
    \begin{tabular}{c|cccccccccc}
    \toprule
    compared & \multicolumn{2}{c}{BNP} & \multicolumn{2}{c}{PPD} & \multicolumn{2}{c}{Bussiness1} & \multicolumn{2}{c}{Bussiness2} & \multicolumn{2}{c}{average} \\
    methods & AUC   & KS    & AUC   & KS    & AUC   & KS    & AUC   & KS    & AUC   & KS \\
    \hline
    LR    & 73.32  & 34.19  & 74.56  & 37.56  & 72.36  & 34.21  & 78.16  & 44.73  & 74.60  & 37.67  \\
    DNN   & 74.08  & 36.28  & 75.81  & 39.65  & 74.60  & 37.11  & 79.68  & 49.33  & 76.04  & 40.59  \\
    GBDT  & 74.46  & 37.24  & 75.37  & 38.79  & 74.89  & 37.16  & 79.54  & 48.54  & 76.07  & 40.43  \\
    CrossNet & 73.43  & 34.47  & 75.17  & 38.36  & 73.94  & 36.76  & 79.07  & 47.74  & 75.40  & 39.33  \\
    CIN   & 73.64  & 35.03  & 75.59  & 39.49  & 74.06  & 36.59  & 79.38  & 48.55  & 75.67  & 39.92  \\
    Self-Attention & 73.97  & 35.89  & 76.12  & 40.72  & 74.39  & 37.06  & 79.51  & 48.71  & 76.00  & 40.60  \\
    FM    & 73.39  & 34.39  & 74.69  & 37.26  & 72.54  & 34.32  & 78.92  & 46.39  & 74.89  & 38.09  \\
    HOFM  & 73.46  & 34.42  & 74.82  & 37.59  & 72.78  & 34.78  & 79.21  & 48.29  & 75.07  & 38.77  \\
    AutoFM & 73.28  & 33.41  & 74.88  & 37.61  & 72.69  & 35.02  & 78.84  & 46.05  & 74.92  & 38.02  \\
    GA$^2$M  & 73.19  & 32.38  & 73.57  & 37.18  & 72.49  & 34.51  & 78.33  & 45.67  & 74.40  & 37.44  \\
    GBDT+LR & 74.23  & 36.66  & 75.56  & 39.11  & 74.67  & 36.92  & 79.47  & 48.69  & 75.98  & 40.35  \\
    AutoCross+LR & 74.73  & 37.64  & 76.06  & 40.81  & 74.81  & 37.01  & 79.39  & 48.75  & 76.25  & 41.05  \\
    DNN2LR & $\ \ $\textbf{75.48}$^*$  & $\ \ $\textbf{38.78}$^*$  & $\ \ $\textbf{76.68}$^*$  & $\ \ $\textbf{41.96}$^*$  & $\ \ $\textbf{75.39}$^*$  & $\ \ $\textbf{38.19}$^*$  & $\ \ $\textbf{79.91}$^*$  & $\ \ $\textbf{50.24}$^*$  & $\ \ $\textbf{76.87}$^*$  & $\ \ $\textbf{42.29}$^*$  \\
    \bottomrule
    \end{tabular}%
  \label{tab:performance}%
\end{table*}%

For LR, GA$^2$M \cite{lou2013accurate} and LR models used in GBDT+LR \cite{he2014practical}, AutoCross+LR \cite{yuanfei2019autocross} and DNN2LR, we tune the learning rate in the range of $\{0.0001,0.001,$ $0.01,0.1,1.0\}$, and the l2 regularization in the range of $\{0.0001,0.001,$ $0.01,0.1,1.0\}$.

For DNN and the DNN model used in DNN2LR, we set the dimensionality of feature embeddings as $10$, the learning rate as $0.001$, and the l2 regularization as $0.0001$.
We use Adam for optimization, use the commonly-applied ReLu as activation function, and tune batch size in the range of $\{256,512,1024,4096\}$.
The DNN model is a multi-layer fully-connected DNN, whose hidden components in deep layers are set to $[400, 100]$.

For FM-based methods, i.e., FM \cite{rendle2010factorization}, HOFM \cite{blondel2016higher} and AutoFM \cite{liu2020autofis}, we set the dimensionality of feature embeddings as $10$, the learning rate as $0.001$, the l2 regularization as $0.0001$, and tune batch size in the range of $\{256,512,1024,4096\}$.
We generate up to forth-order cross features with HOFM.
The efficiency of AutoFM is low when there are too many input feature fields and desired interaction order is high.
Thus, for efficiency, with AutoFM, we generate up to second-order cross features on default prediction datasets with hundreds of feature fields used in our experiments.

For deep learning-based crossing methods, i.e., CrossNet \cite{wang2017deep}, CIN \cite{lian2018xdeepfm} and Self-Attention \cite{song2019autoint,vaswani2017attention}, we set the dimensionality of feature embeddings as $10$, the learning rate as $0.001$, and the l2 regularization as $0.0001$.
We use Adam for optimization, use the commonly-applied ReLu as activation function, and tune batch size in the range of $\{256,512,1024,4096\}$.
For these three methods, we have $3$ layers of feature crossing, which means conducting up to forth-order crossing.
For Self-Attention, the number of hidden units and the number of attention heads are set to $32$ and $2$ respectively.

For GBDT and GBDT+LR \cite{he2014practical}, we set the number of trees to $1000$, with $10$-round early-stopping.
The max depth in the tree is tuned in the range of $\{3,4,5,6,7,8,9,10\}$.

For AutoCross \cite{yuanfei2019autocross}, we divide the dataset into $\sum\nolimits_{i = 0}^{\left\lceil {\mathop {\log }\nolimits_2 t} \right\rceil  - 1} {\mathop 2\nolimits^i }$ batches of data for evaluating the contribution of each candidate cross feature field, where $t$ is the size of candidate set in AutoCross.

For our proposed DNN2LR approach, the quantile threshold $\eta$ for filtering feasible features in the interpretation inconsistency matrix is tuned in the range of $\{10\%,5\%,1\%,0.5\%,0.1\%\}$, and the size $\varepsilon$ of candidate set is tuned in the range of $\{ n, 2n, 3n, 4n, 5n \}$, where $n$ is the size of the original feature fields.

Moreover, our experiments are conducted on a machine with Intel(R) Xeon(R) CPU (E5-26p30 v4 @ 2.20GHz, 24 cores) and 128G memory.
For feature preprocessing, we apply multi-granularity discretization \cite{yuanfei2019autocross} on numerical features.
The granularities in this method are set as $\{10, 100, 1000\}$, and we only keep the best granularity for each numerical feature field.
We evaluate the performances in terms of two commonly-used metrics for credit scoring: the AUC (Area Under the ROC Curve) metric\footnote{\url{https://en.wikipedia.org/wiki/Area_under_the_curve_(pharmacokinetics)}} and the KS (Kolmogorov-Smirnov) metric\footnote{\url{https://en.wikipedia.org/wiki/Kolmogorov\%E2\%80\%93Smirnov_test}}.
AUC measures the overall ranking performance.
KS measures the largest difference between true positive rate and false positive rate on the ROC curve, which can be a reference of the threshold for loan approval.

\subsection{Performance Comparison} \label{sec:performance}

Tab. \ref{tab:performance} shows the performances comparison among different methods.
We can observe that, in most cases, different feature crossing methods can achieve somehow improvements on the performances of LR.
Among deep learning-based cross methods, Self-Attention has good performances, and can slightly outperform DNN and GBDT.
FM-based methods, i.e., FM, HOFM and AutoFM, does not achieve good performances.
This may indicate that, product-based similarity calculation in FM-based methods is not suitable for interactions in credit evaluation tasks.
GA$^2$M performs poor, for it is hard to optimize models with too many candidate cross feature fields on credit scoring datasets with hundreds of input feature fields.
GBDT+LR has relatively good performances.
Obviously, AutoCross+LR is the best one among baseline methods.
However, DNN2LR performs even beeter than AutoCross+LR.
This is because, on credit scoring datasets with hundreds of feature fields, data for training the field-wise LR model of each candidate cross feature field will be too little to produce reliable results in AutoCross.
These experimental results show DNN2LR outperforms conventional classifiers, as well as competitive feature crossing methods on the task of credit scoring.
Moreover, the improvements achieved by DNN2LR are similar between public datasets and business datasets.

\begin{table}[t]
  \centering
    \small
  \caption{Performance comparison between DNN2LR with and without beam search in terms of AUC (\%).}
    \begin{tabular}{c|cccc}
    \toprule
    beam search & BNP   & PPD   & Bussiness1 & Bussiness2 \\
    \hline
    w/    & 75.51  & 76.72  & 75.39  & 79.96  \\
    w/o   & 75.48  & 76.68  & 75.39  & 79.91  \\
    \bottomrule
    \end{tabular}%
  \label{tab:beam}%
\end{table}%

\begin{figure}
\centering
\includegraphics[width=0.32\textwidth]{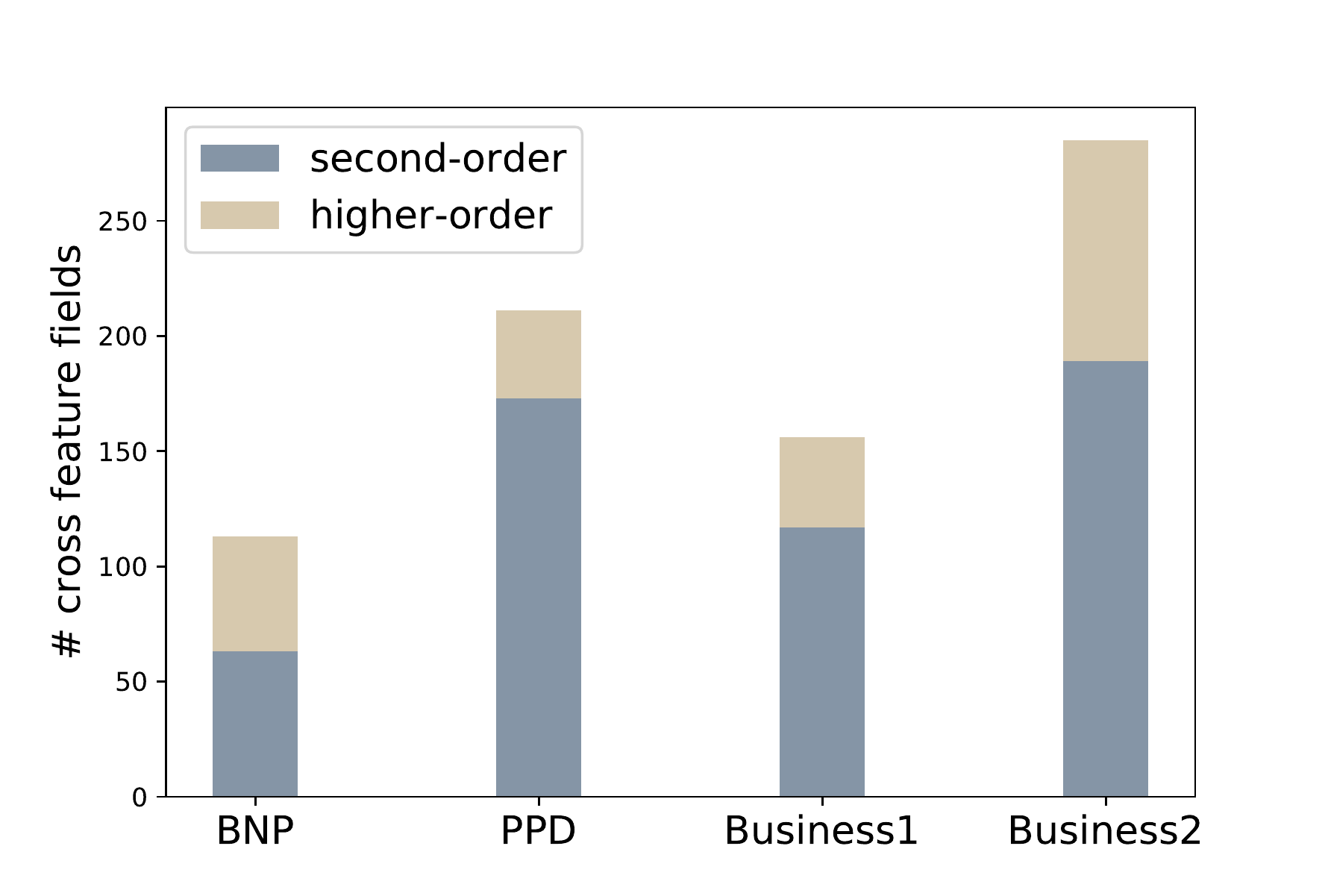}
\caption{The count of second-order and higher-order cross feature fields generated by DNN2LR on each dataset.}
\label{fig:count}
\end{figure}

As discussed in Sec. \ref{sec:searching}, we can incorporate beam search \cite{cohen2019empirical} in Alg. \ref{alg:searching}.
To investigate this, we show performance comparison between DNN2LR with and without beam search in Tab. \ref{tab:beam}.
To be noted, this experiment is done when the DNN is trained and fixed.
With beam search, we keep the best $3$ sets of cross feature fields during each step in Alg. \ref{alg:searching}, and use the set with best overall performance as the final set.
It is clear that, with or without beam search, the performances are close.
Accordingly, it is not necessary to incorporate beam search in the searching procedure of DNN2LR.

Furthermore, as shown in Fig. \ref{fig:count}, we illustrate the count of both second- and higher-order cross feature fields generated by DNN2LR on each dataset.
Roughly speaking, more original feature fields in the dataset, more cross feature fields we need to generate.

\subsection{Running Time Comparison}

\begin{figure}
\centering
\includegraphics[width=0.42\textwidth]{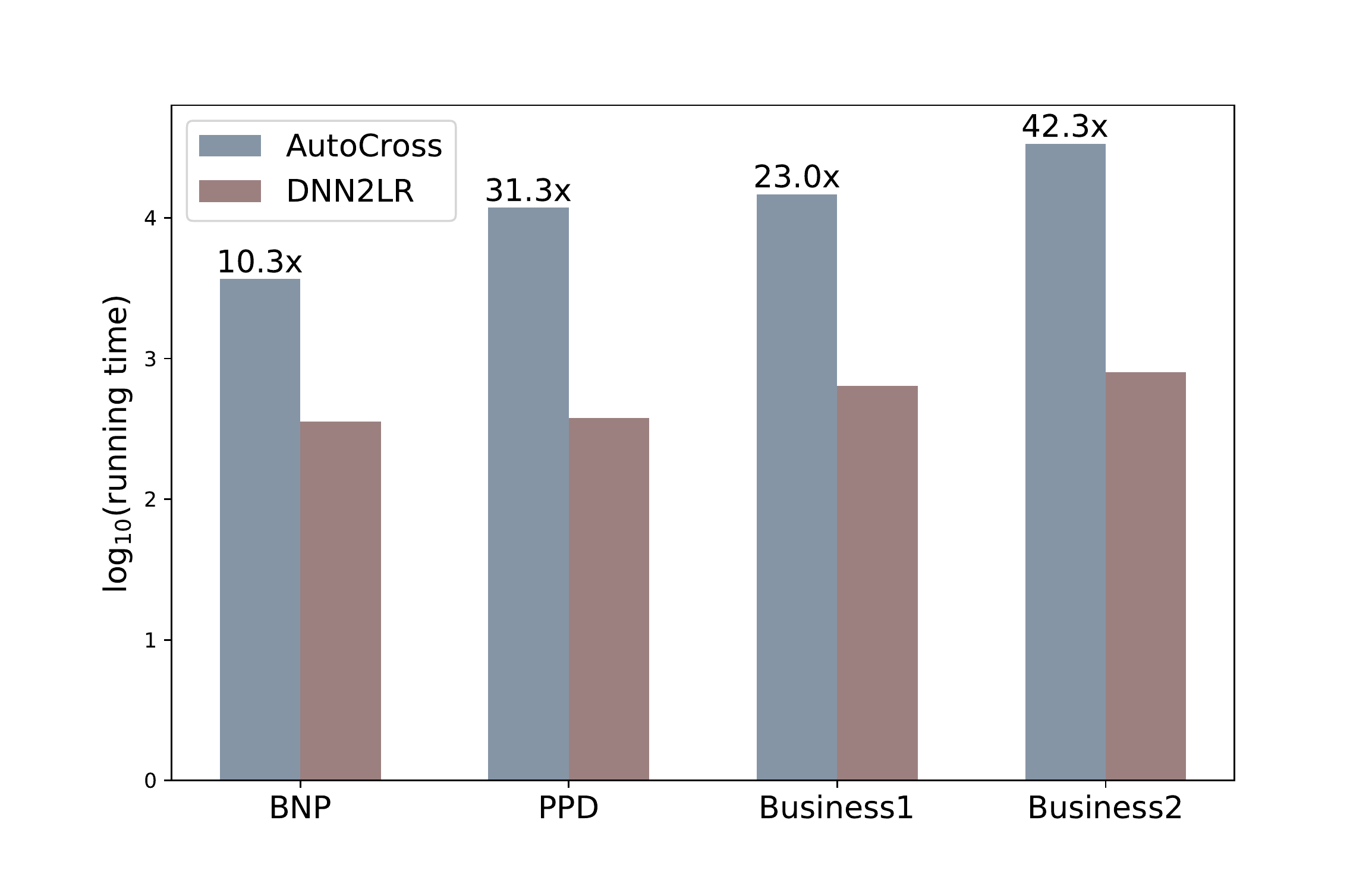}
\caption{Running time comparison between AutoCross and DNN2LR. We report results in terms of seconds after $log_{10}$ calculation. We also show the speedup ratio on each dataset.}
\label{fig:speed}
\end{figure}

For the efficiency comparison, we involve the best baseline method AutoCross, according to Tab. \ref{tab:performance}.
Fig. \ref{fig:speed} shows the running time comparison between AutoCross and DNN2LR.
DNN2LR accelerates AutoCross by about $10\times$, $31\times$, $23\times$ and $42\times$ on BNP, PPD, Business1 and Business2 respectively.
According to Tab. \ref{tab:data}, there are $100$ to $300$ feature fields in these datasets.
It is clear that, the speedup ratios are significant, and get larger when we have more input feature fields in the dataset.
These results strongly illustrate the high efficiency of DNN2LR for feature crossing on credit scoring datasets with hundreds of feature fields.

\begin{figure}
\centering
\includegraphics[width=0.42\textwidth]{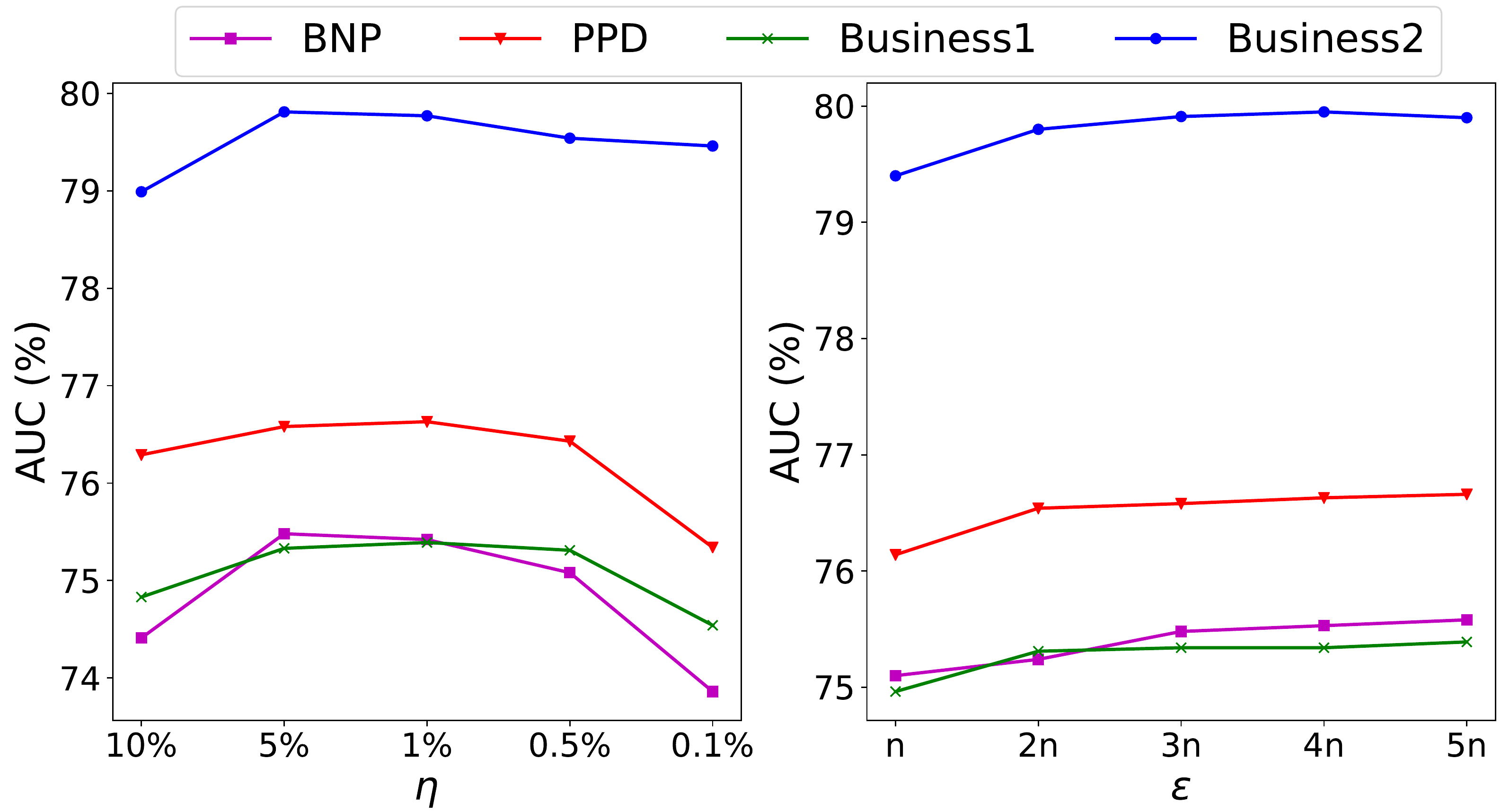}
\caption{Performances of DNN2LR with varying hyper-parameters: (1) the left part shows the impact of the quantile threshold $\eta$ for filtering feasible features in the interpretation inconsistency matrix; (2) the right part shows the impact of the size $\varepsilon$ of candidate set, where $n$ is the number of original feature fields in each dataset.}
\label{fig:params}
\end{figure}

\subsection{Hyper-parameter Study} \label{sec:params}

As shown in Fig. \ref{fig:params}, we illustrate the sensitivity of DNN2LR to hyper-parameters.
First, we investigate the performances of DNN2LR with varying quantile thresholds $\eta$ for filtering feasible features in the interpretation inconsistency matrix.
According to the curves in Fig. \ref{fig:params}, the performances of DNN2LR are relatively stable, especially in the range of $\{5\%,1\%,0.5\%\}$.
Then, we investigate the performances of DNN2LR with varying sizes $\varepsilon$ of candidate set.
As discussed in Sec. \ref{sec:performance}, we need to generate more cross feature fields when we have more original feature fields.
Thus, we make the size of candidate set related to the number $n$ of original feature fields.
We can observe that, the curves are stable when $\varepsilon \ge 2n$.
And when we have larger $\varepsilon$, the performances of DNN2LR slightly increase.
According to our observations, we set $\eta = 5\%$ and $\varepsilon = 3n$ as the optimal hyper-parameters in our experiments.
This makes the candidate size extremely small, comparing to the set of possible cross feature fields on credit scoring datasets with hundreds of feature fields.

\begin{figure*}
	\centering
	\subfigure[\emph{In other words}, it's just another sports study.]{
		\begin{minipage}[b]{0.32\textwidth}
			\includegraphics[width=1\textwidth]{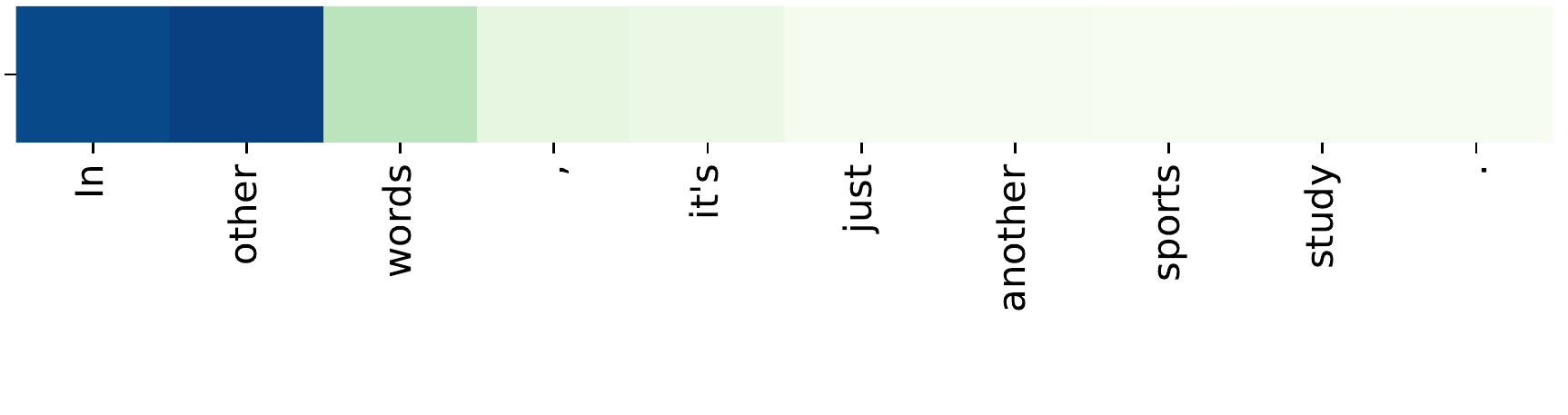}
		\end{minipage}
	}
	\subfigure[\emph{Can't} seem to \emph{get anywhere near} the story's center.]{
		\begin{minipage}[b]{0.32\textwidth}
			\includegraphics[width=1\textwidth]{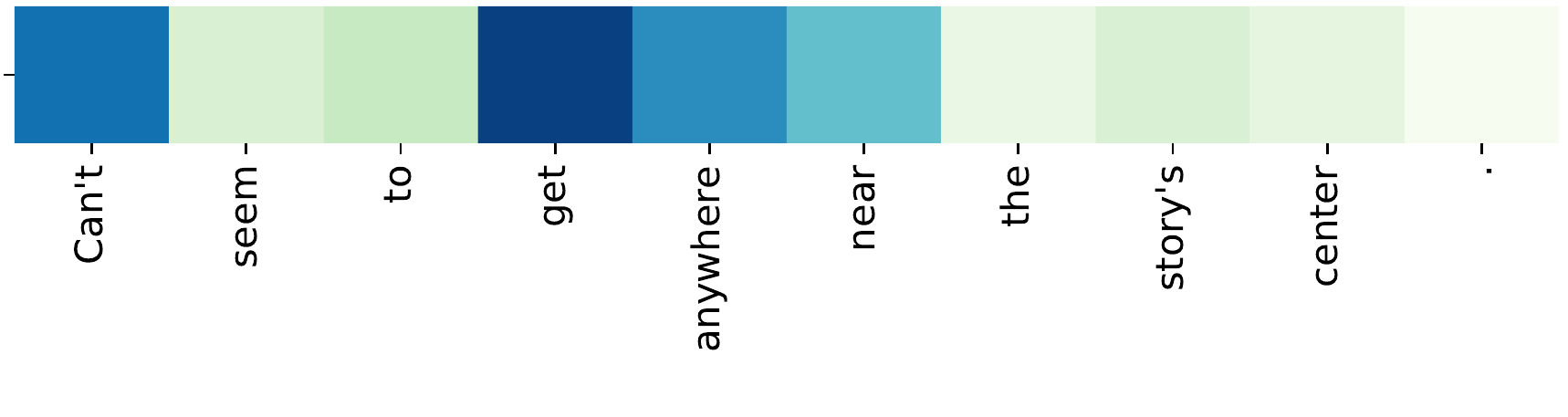}
		\end{minipage}
	}
	\subfigure[Makes good B movies, and the Scorpion King more than ably \emph{meets those standards}.]{
		\begin{minipage}[b]{0.32\textwidth}
			\includegraphics[width=1\textwidth]{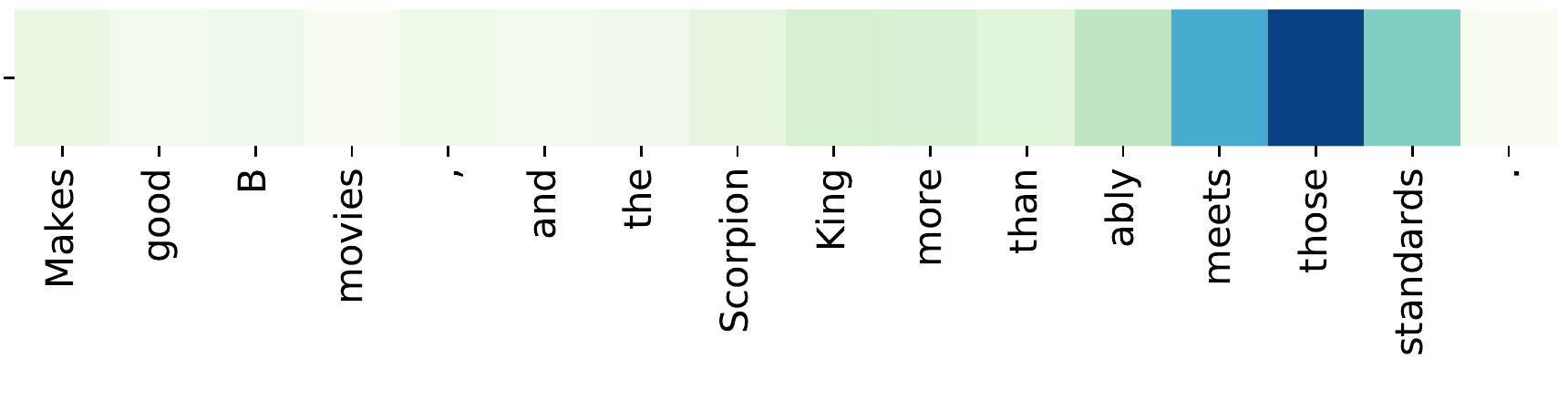}
		\end{minipage}
	}
	\subfigure[The only reason you should see this movie is if you have a case of masochism and an \emph{hour and a half to blow}.]{
		\begin{minipage}[b]{0.32\textwidth}
			\includegraphics[width=1\textwidth]{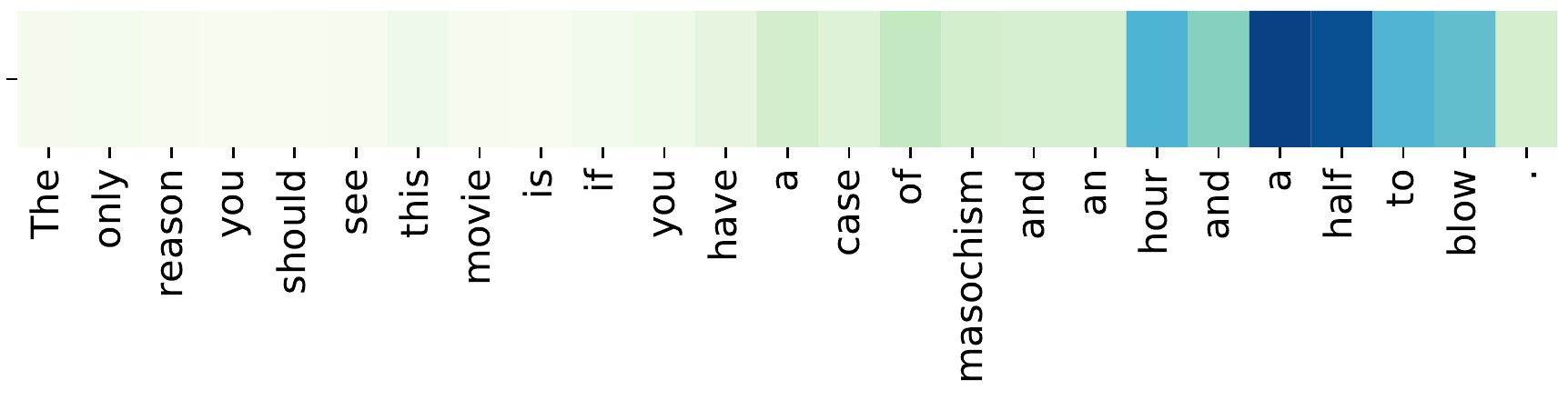}
		\end{minipage}
	}
	\subfigure[Roman Polanski directs the pianist like a surgeon mends a broken heart; very meticulously but \emph{without any passion}.]{
		\begin{minipage}[b]{0.32\textwidth}
			\includegraphics[width=1\textwidth]{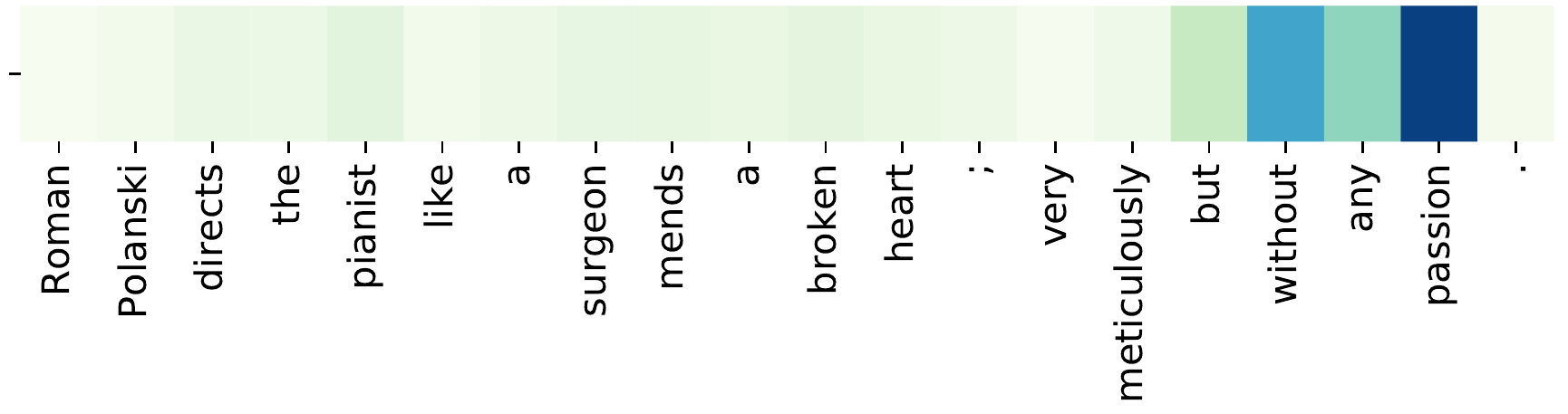}
		\end{minipage}
	}
	\subfigure[Another one of those estrogen overdose movies, except that the writing, acting and character development are \emph{a lot better}.]{
		\begin{minipage}[b]{0.32\textwidth}
			\includegraphics[width=1\textwidth]{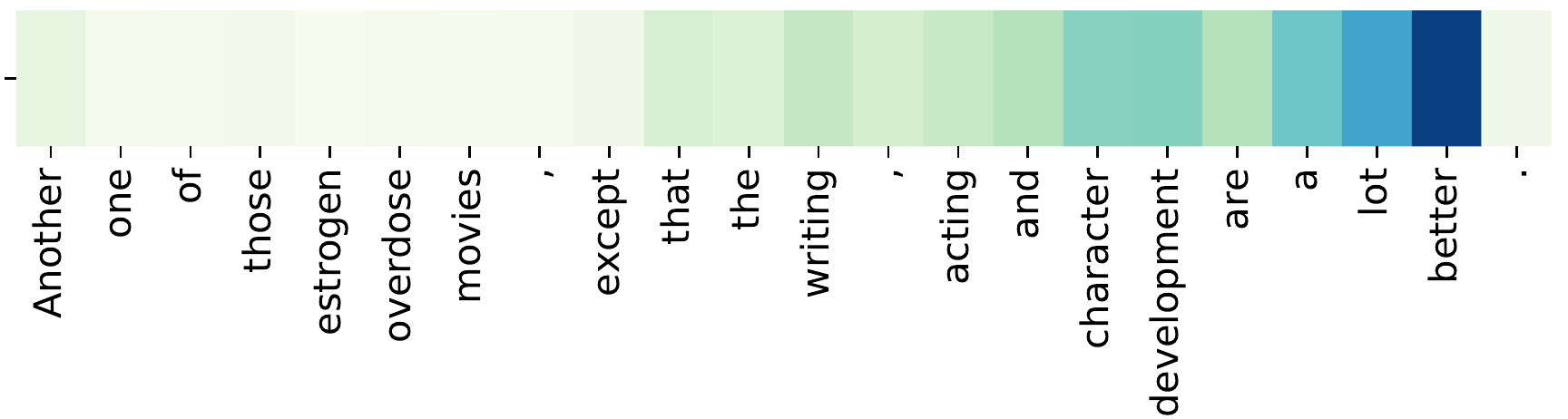}
		\end{minipage}
	}
	\subfigure[Even at an hour and twenty minutes, it's too long and \emph{it goes nowhere}.]{
		\begin{minipage}[b]{0.32\textwidth}
			\includegraphics[width=1\textwidth]{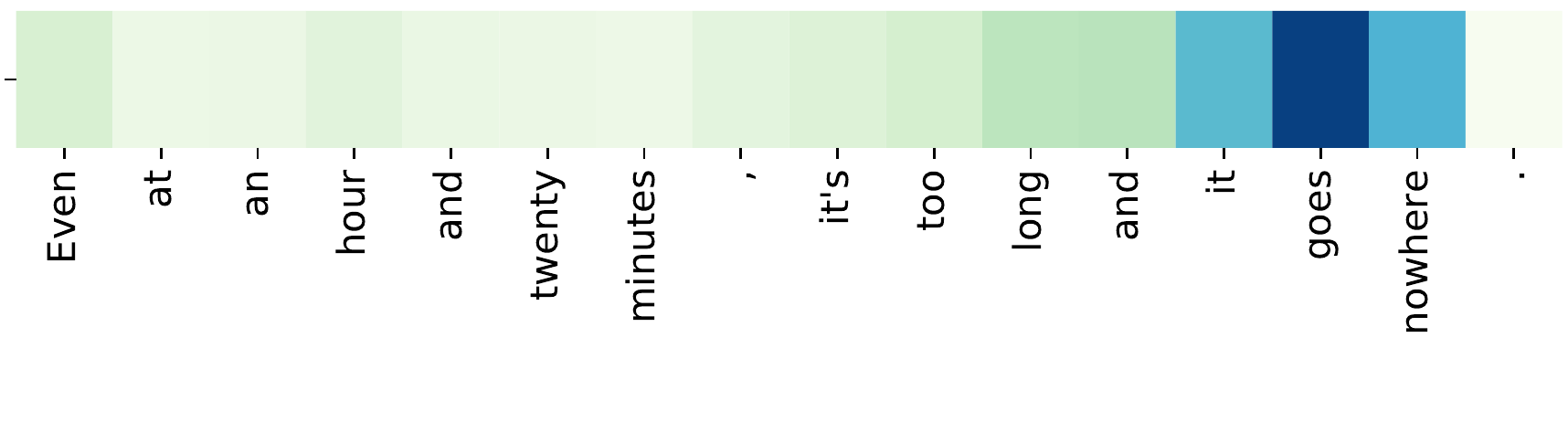}
		\end{minipage}
	}
	\subfigure[Maybe it's asking too much, but if a movie is truly trying to inspire me, I \emph{want} a \emph{little more than this}.]{
		\begin{minipage}[b]{0.32\textwidth}
			\includegraphics[width=1\textwidth]{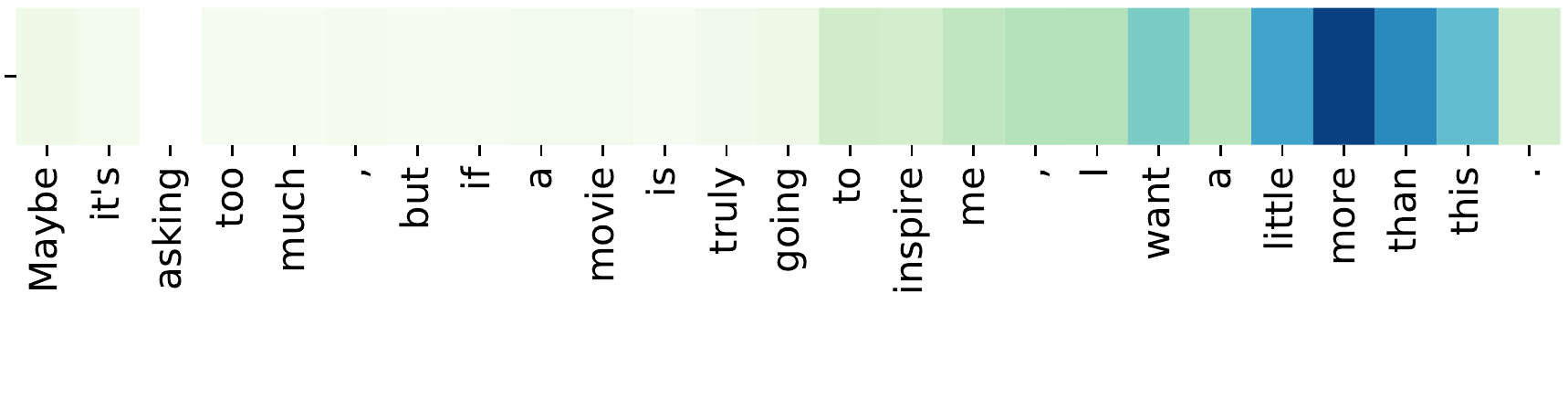}
		\end{minipage}
	}
	\subfigure[I'll bet that the video game is a lot \emph{more fun than the film}.]{
		\begin{minipage}[b]{0.32\textwidth}
			\includegraphics[width=1\textwidth]{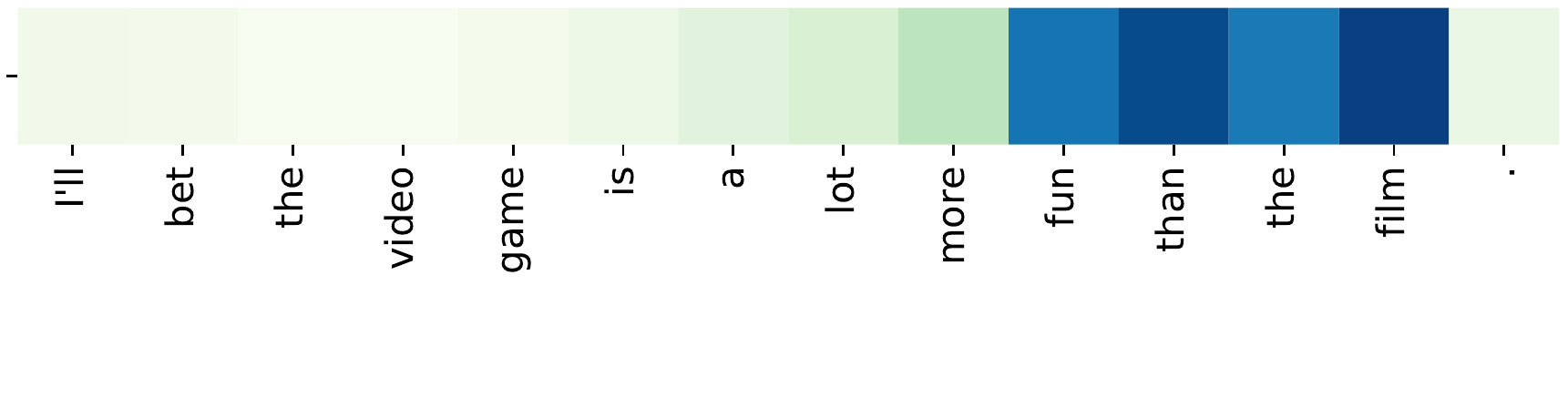}
		\end{minipage}
	}
	\caption{Interpretation inconsistency values of words in the BiLSTM model on the MR dataset. The values are normalized in each sentence. Deeper the square, larger the corresponding value. Better viewed in color.}
	\label{fig:vis}
\end{figure*}

\begin{table}[t]
 \centering
 \small
 \caption{Average interpretation inconsistency values of feature fields $\alpha$, $\beta$ and $\lambda$ in several numerical formulations.}
    \begin{tabular}{c|ccc}
    \toprule
    formulation & $\alpha$ & $\beta$ & $\lambda$ \\
    \hline
    $\alpha  \times \beta  + \lambda$ & 0.0159 & 0.0162 & 0.0034 \\
    $\alpha  \times \beta  + \lambda^2$ & 0.0231 & 0.0149 & 0.0056 \\
    $\alpha  \times \beta^2  + \lambda^3$ & 0.0137 & 0.0193 & 0.0039 \\
    $\alpha^3  \times \beta^2  + \lambda^3$ & 0.0529 & 0.0527 & 0.0084 \\
    $\alpha /\left( {1 + 10 \times \beta } \right) + 1/\left( {0.1 + \lambda } \right)$ & 1.2300 & 0.7514 & 0.2281 \\
    $\alpha /\left( {0.1 + \beta } \right) + 1/\left( {0.1 + \lambda } \right)$ & 2.7073 & 1.3112 & 0.4556 \\
    $\left( {1 + \alpha^2 } \right) /\left( {0.5 + \beta } \right) + \lambda^2$ & 0.0888 & 0.0743 & 0.0113 \\
    $\ln \left( {1 + 10 \times \alpha  \times \beta } \right) + \lambda$ & 0.0495 & 0.0523 & 0.0027 \\
    $\ln \left( {1 + 10 \times \alpha  \times \beta } \right) + \lambda^2$ & 0.0591 & 0.0568 & 0.0081 \\
    $\ln \left( {1 + 10 \times \alpha  \times \beta } \right) + e^{1 + \lambda }$ & 0.0802 & 0.1255 & 0.0156 \\
    $\ln \left( {1 + \alpha } \right) \times \beta + \lambda^2$ & 0.0131 & 0.0141 & 0.0046 \\
    $e^{1+\alpha} \times \beta^2 + \lambda$ & 0.5625 & 0.6897 & 0.1782 \\
    $e^{1+\alpha \times \beta} + \lambda^2$ & 0.5151 & 0.4948 & 0.0667 \\
    $e^{1+\alpha \times \beta} + 10 \times \lambda$ & 0.2828 & 0.2931 & 0.0726 \\
    \bottomrule
    \end{tabular}%
 \label{tab:formulation}%
\end{table}%

\subsection{Interpretation Inconsistency} \label{sec:vis}

To investigate the validity of interpretation inconsistency for feature crossing, we conduct some empirical experiments.
We conduct experiments on some mathematic formulations, which are shown in Tab. \ref{tab:formulation}.
In each formulation, there are three feature fields: $\alpha$, $\beta$ and $\lambda$, where $\alpha$ and $\beta$ are interacted, while $\lambda$ has no interaction with the other two fields.
The interactions between $\alpha$ and $\beta$ cover different forms, and the calculation of $\lambda$ also varies from simple to complex.
For each formulation, we randomly sample features in $[0.0, 1.0]$, and keep two decimal places.
We perform DNN on each formulation, and calculate the average interpretation inconsistency values of the three feature fields according to Eq. (\ref{equation:interpretation_inconsistency}), as shown in Tab. \ref{tab:formulation}.
To be noted, we do not perform feature discretization in this experiment.
It is clear that, in each formulation, the average interpretation inconsistency values of $\alpha$ and $\beta$ are much larger than that of $\lambda$.
This observation clearly confirms to the formulations.

Furthermore, we conduct experiments on sentence classification.
We train BiLSTM on the MR dataset\footnote{\url{http://www.cs.cornell.edu/people/pabo/movie-review-data/}}.
There are $5331$ and $5331$ positive and negative movie reviews respectively.
Values of interpretation inconsistency in several example sentences are shown in Fig. \ref{fig:vis}.
We achieve two observations about words with large values of interpretation inconsistency.
First, words that form phrases have large values of interpretation inconsistency, e.g., ``in other words," ``can't ... get anywhere near," ``meets those standards," ``an hour and a half to blow" and ``it goes nowhere."
Second, words with specific intensities or orientations have large values of interpretation inconsistency, e.g., ``a lot more fun than the film," ``a lot better" and ``a little more than this."
These observations confirm to the feature interactions among words in sentences.
These experiments suggest that it is valid to learn feature crossing from interpretation inconsistency in DNN.

\begin{table}[t]
  \centering
  \scriptsize
  \caption{Top-$20$ cross feature fields generated by DNN2LR.}
    \begin{tabular}{cc}
    \toprule
    BNP   & PPD \\
    \hline
    (v22,v24,v40,v79) & (ThirdParty\_Info\_Period3\_1,UserInfo\_16) \\
    (v50,v110) & (ListingInfo,Education\_Info1) \\
    (v30,v50) & (UserInfo\_14,UserInfo\_16) \\
    (v22,v50,v113) & (ThirdParty\_Info\_Period2\_8,UserInfo\_16) \\
    (v24,v50) & (ThirdParty\_Info\_Period3\_2,UserInfo\_16) \\
    (v22,v24,v79,v113) & (UserInfo\_16,ListingInfo) \\
    (v47,v50) & (WeblogInfo\_2,UserInfo\_14) \\
    (v56,v66) & (ThirdParty\_Info\_Period1\_10,UserInfo\_16) \\
    (v47,v50,v110) & (ThirdParty\_Info\_Period3\_15,UserInfo\_14) \\
    (v10,v22,v40,v113) & (ListingInfo,UserInfo\_7,UserInfo\_19) \\
    (v21,v50) & (UserInfo\_16,UserInfo\_3,Education\_Info1) \\
    (v50,v56,v113) & (WeblogInfo\_20,ThirdParty\_Info\_Period1\_10) \\
    (v40,v50) & (ListingInfo,WeblogInfo\_5) \\
    (v24,v66) & (UserInfo\_14,ThirdParty\_Info\_Period2\_15) \\
    (v10,v50) & (ThirdParty\_Info\_Period4\_6,ThirdParty\_Info\_Period2\_6,ListingInfo) \\
    (v22,v30,v113) & (WeblogInfo\_15,WeblogInfo\_20) \\
    (v47,v50,v56) & (ThirdParty\_Info\_Period3\_6,UserInfo\_15) \\
    (v50,v113) & (UserInfo\_16,WeblogInfo\_20) \\
    (v50,v79) & (WeblogInfo\_15,SocialNetwork\_13) \\
    (v40,v56,v110,v113) & (WeblogInfo\_2,ThirdParty\_Info\_Period4\_15) \\
    \bottomrule
    \end{tabular}%
  \label{tab:importance}%
\end{table}%

\subsection{Important Cross Feature Fields}

In Tab. \ref{tab:importance}, we illustrate top-$20$ cross feature fields generated by DNN2LR on public datasets, i.e., BNP and PPD.
Once the meanings of original feature fields are known, we can obtain the meanings of cross feature fields.
The final model generated by DNN2LR is a LR model empowered with cross features, which is a white-box model with high accuracy and global interpretability simultaneously.

\section{Conclusion}

In this paper, we focus on the automatic feature crossing task for credit scoring, in which we usually face hundreds of feature fields.
We observe that the local interpretations of a specific feature are usually inconsistent in different samples, and accordingly define interpretation inconsistency in DNN.
We show that, the interpretation inconsistency in DNN can help us to find cross features.
Then, we propose a novel automatic feature crossing method called DNN2LR for credit scoring.
DNN2LR constructs an accurate and compact candidate set of cross feature fields based on interpretation inconsistency, and search for the final set of cross feature fields to train a final LR model.
Thus, with DNN2LR, we can obtain a white-box model, i.e., a LR model empowered with cross features.
Extensive experiments have been conducted on real-world credit scoring datasets with large numbers of feature fields.
Experimental results demonstrates that, DNN2LR outperforms several conventional classifiers, as well as several automatic feature crossing methods.
Moreover, comparing with the state-of-the-art method AutoCross, DNN2LR can significantly accelerate the speed by about $10\times$ to $40\times$.
In a word, DNN2LR is able to efficiently produce accurate white-box models for the task of credit scoring.


\bibliographystyle{ACM-Reference-Format}
\bibliography{ref}

\end{document}